\documentclass{article}
\usepackage{iclr2026_conference,times}

\usepackage[utf8]{inputenc}
\usepackage[T1]{fontenc}
\usepackage{hyperref}
\usepackage{url}
\usepackage{booktabs}
\usepackage{amsfonts}
\usepackage{amsmath}
\usepackage{nicefrac}
\usepackage{microtype}
\usepackage{graphicx}
\usepackage{xcolor}
\usepackage{multirow}
\usepackage{subcaption}
\usepackage{wrapfig}
\usepackage{algorithm}
\usepackage{algorithmic}

\newif\ifshowTodos
\showTodosfalse 

\title{A Lightweight Library for Energy-Based Joint-Embedding Predictive Architectures}

\author{\textbf{Basile Terver$^{1,2}$, Randall Balestriero$^1$, Megi Dervishi$^1$, David Fan$^1$,}\\
\textbf{Quentin Garrido$^1$, Tushar Nagarajan$^1$, Koustuv Sinha$^1$, Wancong Zhang$^1$,}\\
\textbf{Mike Rabbat$^{1}$, Yann LeCun$^{1,3,\dag}$, Amir Bar$^{1,\dag}$}
\\
$^1$Meta FAIR \quad $^2$INRIA \quad $^3$New York University\\
$^\dag$Equal Contribution
}

\iclrfinalcopy

\begin{document}

\maketitle
\fancyhead[L]{ICLR 2026, the 2nd Workshop on World Models}

\begin{abstract}
We present \textbf{EB-JEPA}, an open-source library for learning representations and world models using Joint-Embedding Predictive Architectures (JEPAs).
JEPAs learn to predict in representation space rather than pixel space, avoiding the pitfalls of generative modeling while capturing semantically meaningful features suitable for downstream tasks.
Our library provides modular, self-contained implementations that illustrate how representation learning techniques developed for image-level self-supervised learning can transfer to video, where temporal dynamics add complexity, and ultimately to action-conditioned world models, where the model must additionally learn to predict the effects of control inputs.
Each example is designed for single-GPU training within a few hours, making energy-based self-supervised learning accessible for research and education.
We provide ablations of JEA components on CIFAR-10. Probing these representations yields 91\% accuracy, indicating that the model learns useful features. Extending to video, we include a multi-step prediction example on Moving MNIST that demonstrates how the same principles scale to temporal modeling. Finally, we show how these representations can drive action-conditioned world models, achieving a 97\% planning success rate on the Two Rooms navigation task.
Comprehensive ablations reveal the critical importance of each regularization component for preventing representation collapse.
\footnote{Code is available at \url{https://github.com/facebookresearch/eb_jepa}.}
\end{abstract}

\section{Introduction}

The idea that intelligent systems should learn internal models of their environment has deep roots in cognitive science, from early theories of mental models~\citep{craik1967nature} to predictive coding accounts of perception~\citep{Rao99} and learned world models for planning~\citep{Dyna1991,Schmidhuber1990}.
Recent advances in video generation~\citep{brooks2024video,blattmann2023stablevideodiffusionscaling} and interactive world simulators~\citep{bruce2024genie,genie2} have shown impressive results, but those face fundamental challenges: they must model all pixels, {\em including task-irrelevant details}, thereby requiring substantial computational resources~\citep{ReconstructionUninformativebalestriero24b}.
Joint-Embedding Predictive Architectures (JEPAs)~\citep{PathAMI,IJEPA,VJEPA} offer an alternative paradigm.
Rather than reconstructing observations in pixel space, JEPAs learn to predict in a learned representation space, focusing computational effort on semantically meaningful features.

\begin{figure}[t]
\vspace{-2em}
\centering
\includegraphics[width=1\textwidth, trim= 10pt 4pt 0pt 4pt]{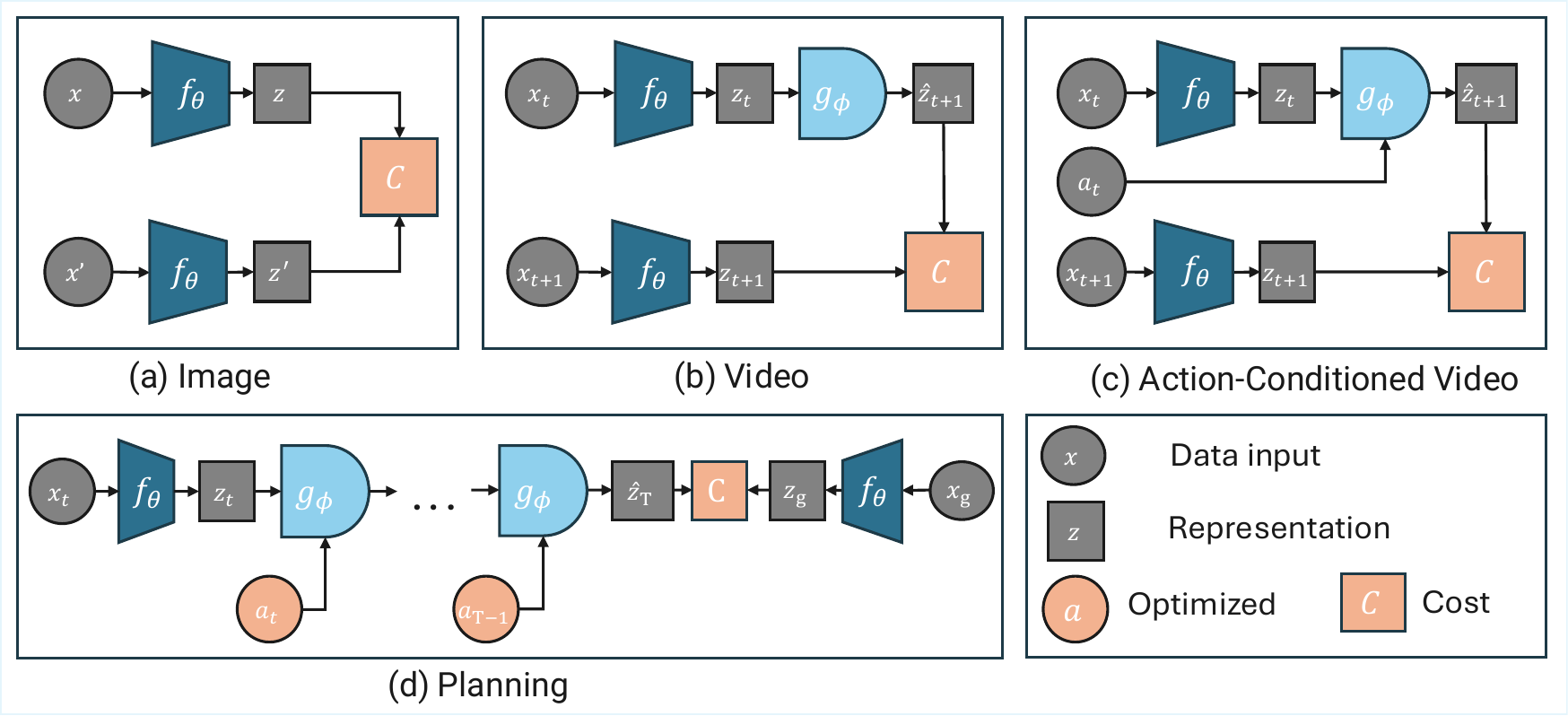}
\caption{\textbf{EB-JEPA} is a modular code base and tutorial, providing self-contained implementations of Joint-Embedding Predictive Architecture for (a) self-supervised image representation learning, (b) video prediction in latent space, and (c) action-conditioned world models that enable goal-directed planning (d).}
\label{fig:teaser}
\vspace{-1em}
\end{figure}

JEPA builds on a rich history of self-supervised representation learning~\citep{SimCLR,MoCo,BYOL,BarlowTwins,SimSiam}.
JEPAs have demonstrated strong performance for visual representation learning~\citep{IJEPA} and have been extended to video understanding~\citep{VJEPA} and world modeling for planning~\citep{VJEPA2,PLDM,DINO-WM,terver2026JEPAWMs}.
Despite this growing body of work, accessible implementations that bridge theoretical principles and practical application remain scarce.
Production-scale implementations are designed for large-scale training and are challenging to navigate.
World model implementations like DINO-WM~\citep{DINO-WM} and JEPA-WMs~\citep{terver2026JEPAWMs} enable planning on simple environments but rely on particular setups, e.g., frozen pre-trained encoders. As a result, while JEPAs have shown promise, they still have a high barrier to entry, which we hope to address in this study.
This paper introduces \textbf{EB-JEPA} (Figure~\ref{fig:teaser}), an open-source library that addresses this gap through modular, well-documented implementations of JEPA-based models trainable at small scale with simple, concise code designed for educational purposes and rapid experimentation.
Our contributions are:
\begin{enumerate}
    \item \textbf{Accessible implementations}: Three progressively complex examples (image representation learning, video prediction, and action-conditioned planning), each trainable on a single GPU in a few hours.
    \item \textbf{Modular architecture}: Reusable components (encoders, predictors, regularizers, planners) that can be easily recombined for new applications.
    \item \textbf{Comprehensive evaluation}: Systematic experiments and ablations demonstrating the importance of each component, with practical guidance on hyperparameter selection.
    \item \textbf{Educational resource}: Clear documentation and code structure designed to help researchers understand JEPA principles.
\end{enumerate}

\section{Related Work}

\paragraph{Joint-Embedding methods.}
EB-JEPA builds on the JEPA framework~\citep{IJEPA,VJEPA}, focusing on their subclass using regularization-based collapse prevention~\citep{VICReg,BarlowTwins,balestriero2025lejepaprovablescalableselfsupervised} rather than stop-gradient techniques~\citep{BYOL,SimSiam,DINOV2}.
Recent theoretical work has provided deeper understanding of these methods: \citet{InformationTheoPerspectiveVICREG2023} analyze VICReg from an information-theoretic perspective, while \citet{SLLspectralembeddings2022} show connections between contrastive and non-contrastive methods and spectral embedding.
While I-JEPA and V-JEPA focus on masked prediction within single images or videos, our action-conditioned example extends this to interactive settings where actions determine future states.
Recent work has shown that JEPA-style pretraining leads to emergent understanding of intuitive physics~\citep{garrido2025intuitivephysicsunderstandingemerges}, motivating the use of such architectures for world modeling.
Importantly, JEPAs differ fundamentally from reconstruction-based methods such as MAE~\citep{MAE} and VideoMAE~\citep{VideoMAE,VideoMAEv2}, which predict in pixel space rather than representation space.
\citet{ReconstructionUninformativebalestriero24b} provide theoretical analysis showing that reconstruction-based learning can produce uninformative features for perception, further motivating the joint-embedding paradigm that our library focuses on.

\paragraph{World models for planning.}
Latent world models have been extensively studied for model-based reinforcement learning (RL)~\citep{PlaNet,DreamerV3,TDMPC2}.
Our work is most closely related to PLDM~\citep{PLDM}, IWM~\citep{IWM}, DINO-WM~\citep{DINO-WM}, Navigation World Models~\citep{NWM_Bar_2025_CVPR}, and JEPA-WMs~\citep{terver2026JEPAWMs}, which use joint-embedding objectives for planning.
Unlike these works, we focus on providing accessible, educational implementations rather than state-of-the-art performance on complex benchmarks.

\section{Preliminaries: A Unified JEPA Framework}

Our goal is to train models that map inputs to latent semantic representations useful for perception, planning, and action.
We view this through the lens of \emph{Energy-Based Models} (EBMs)~\citep{LeCunEBMTutorial2006,Hopfield1982}. An EBM defines a scalar energy function $E(x, y)$ measuring compatibility between inputs $x$ and outputs $y$, where low energy indicates high compatibility.
Learning consists of shaping the energy landscape so that correct input-output pairs have lower energy than incorrect ones.

The key challenge in training EBMs is preventing \textit{collapse}: a degenerate solution where the energy is uniformly low for all inputs.
Different strategies address this challenge: contrastive methods push up energy on negative samples~\citep{ContrastiveDivergence2002,SimCLR,MoCo}; stop-gradient and exponential moving average (EMA) techniques break symmetry~\citep{BYOL,SimSiam,IJEPA,VJEPA}; and regularization-based approaches maintain representation diversity without negative samples~\citep{VICReg,BarlowTwins,balestriero2025lejepaprovablescalableselfsupervised}.
Our library focuses on the regularization approach: we instantiate JEPAs with explicit regularization losses to prevent collapse, defining energy as prediction error in representation space.
With the regularizer $\mathcal{R}$ and a given prediction loss $\mathcal{L}_{\text{pred}}$, the JEPA general training objective takes the form
\begin{equation}
\mathcal{L} = \mathcal{L}_{\text{pred}}(g_\phi(z, u), z') + \lambda \mathcal{R}(z),
\label{eq:general_jepa}
\end{equation}
where $z = f_\theta(x)$ is the representation of input $x$, $u=q_\omega(a)$ is optional conditioning information (e.g., robotic controls), $z'$ is the target representation, and $\lambda$ balances prediction and regularization.
This unified framework encompasses three instantiations of increasing complexity, detailed below.

\textbf{(i) Image-JEPA: view invariance.}
Given an image $x$, we create two views $x$ and $x'$ (random crops, color jittering, etc.).
The encoder produces representations $z = f_\theta(x)$ and $z' = f_\theta(x')$.
The objective learns representations invariant to different views, with the energy function
\begin{equation}
\mathcal{L}_{\text{image}} = \|z - z'\|_2^2 + \lambda \mathcal{R}(z, z').
\label{eq:image_jepa_energy}
\end{equation}
Here, the energy directly measures how similar the representations of two views of the same image are. Low energy means the model has learned view-invariant features.

\textbf{(ii) Video-JEPA: temporal prediction.}
We denote a video sequence as $x_{1:T}:=(x_1, \ldots, x_T)$. The encoder produces per-frame representations $z_t = f_\theta(x_{t-w:t})$, where $w$ is the encoder temporal receptive field.
A predictor takes a context of $v+1$ frame representations, where $v$ is the predictor temporal receptive field (see hyperparameter values in Tab.~\ref{tab:video_hyperparams}), and predicts the next representation, yielding the energy
\begin{equation}
\mathcal{L}_{\text{video}} = \sum_{t=1}^{T-1} \|g_\phi(z_{t-v:t}) - z_{t+1}\|_2^2 + \lambda \mathcal{R}(z_{1:T}).
\label{eq:video_jepa_energy}
\end{equation}
The model learns to capture temporal dynamics without access to future frames during prediction.

\textbf{(iii) Action-conditioned video-JEPA (AC-video-JEPA): world modeling.}
Given observation-action sequences $(x_t, a_t)_{t=1}^T$, an action encoder $q_\omega$ maps actions to control representations $u_t = q_\omega(a_{t-w:t})$, and the predictor is conditioned on these representations, yielding the energy
\begin{equation}
\mathcal{L}_{\text{world}} = \sum_{t=1}^{T-1} \|g_\phi(z_{t-v:t}, u_{t-v:t}) - z_{t+1}\|_2^2 + \lambda \mathcal{R}(z_{1:T}, u_{1:T}).
\label{eq:ac_video_jepa_energy}
\end{equation}
This learns a latent dynamics model suitable for planning: given a current state and control representation, predict the next state representation.

\paragraph{A unified energy formulation.}
The three settings above share a common structure. Given an encoder $f_\theta$, a predictor $g_\phi$, and optional conditioning $a$ with conditioning encoder $q_\omega$, we can write a general energy function
\begin{equation}
E(x, x', a) = \mathcal{L}_{\text{pred}}(g_\phi(f_\theta(x), q_\omega(a)), f_\theta(x')).
\label{eq:energy}
\end{equation}
Image-JEPA corresponds to $g_\phi = \text{Id}$ (identity) and no conditioning; video-JEPA uses a temporal predictor without conditioning; AC-video-JEPA includes the full formulation with action conditioning.
This unified view highlights how the same energy-based principle, minimizing prediction error in representation space, underlies all three settings, with complexity increasing as we move from static images to video to action-conditioned dynamics.

\paragraph{Regularization: Preventing Collapse.}
\label{sec:regularization}
The key challenge in training JEPAs is preventing \textit{representation collapse}, where the encoder learns trivial constant representations.
EB-JEPA implements two regularization families.
\textit{VICReg}~\citep{VICReg} prevents collapse through two complementary terms.
The \textit{variance} term ensures each feature dimension has sufficient spread across the batch and reads
\begin{equation}
\mathcal{L}_{\text{var}}(Z) = \frac{1}{D} \sum_{j=1}^{D} \max\left(0, \gamma - \sqrt{\text{Var}(Z_{:,j}) + \epsilon}\right),
\end{equation}
where $Z \in \mathbb{R}^{N \times D}$ is the batch of embeddings, $D$ is the feature dimension, and $\gamma$ is the target standard deviation (typically 1).
The \textit{covariance} term decorrelates feature dimensions to encourage the model to use all available capacity and reads
\begin{equation}
\mathcal{L}_{\text{cov}}(Z) = \frac{1}{D(D-1)} \sum_{i \neq j} [C(Z)]^2_{i,j}, \quad C(Z) = \frac{1}{N-1}(Z - \bar{Z})^\top(Z - \bar{Z}).
\end{equation}
The full VICReg regularizer is $\mathcal{R}_{\text{VICReg}} = \alpha \mathcal{L}_{\text{var}} + \beta \mathcal{L}_{\text{cov}}$.

For image-JEPA and video-JEPA, the regularization losses are computed in a projected space rather than directly on the encoder outputs.
A learned projector $h_\psi$ maps \emph{representations} to \emph{embeddings} $r = h_\psi(z)$ on which the regularizer is computed.
\textit{LeJEPA}~\citep{balestriero2025lejepaprovablescalableselfsupervised} introduces SIGReg, a theoretically grounded alternative regularizer.
It identifies the isotropic Gaussian $\mathcal{N}(0, I)$ as the optimal embedding distribution for minimizing downstream prediction risk.
The SIGReg objective enforces this by testing Gaussianity along random 1D projections $\xi_p \sim \mathcal{N}(0, I)$ and reads
\begin{equation}\label{eq:SIGReg}
\mathcal{R}_{\text{SIGReg}}(Z) = \frac{1}{P}\sum_{p=1}^{P} \mathcal{G}(Z \xi_p),
\end{equation}
where $\mathcal{G}$ is the Epps-Pulley Gaussianity test statistic.
This approach offers a single hyperparameter $\lambda$, linear time/memory complexity, and stability across architectures.

\begin{figure}[t]
\centering
\includegraphics[width=0.9\textwidth]{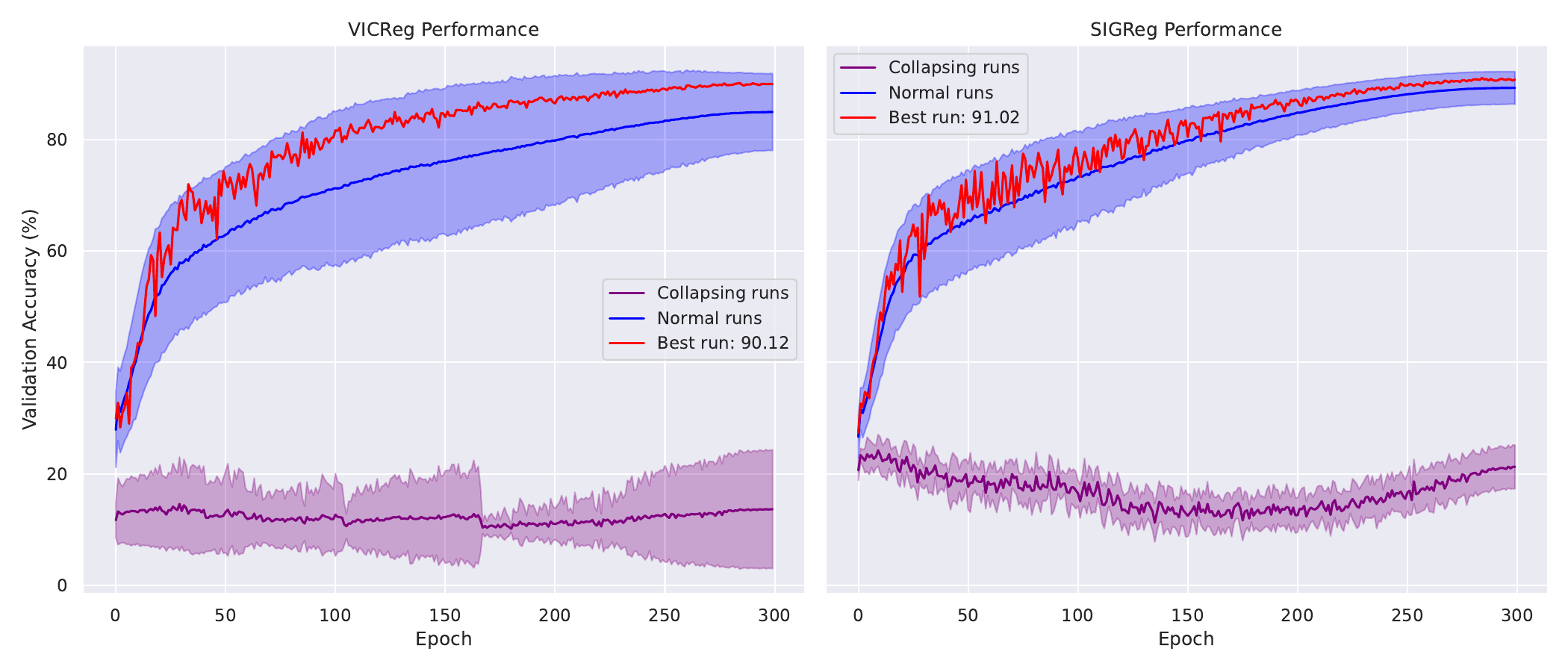}
\caption{Hyperparameter sensitivity comparison between SIGReg and VICReg on CIFAR-10. SIGReg demonstrates greater stability across different hyperparameter configurations, while VICReg achieves similar peak performance but requires more careful tuning.
}
\label{fig:hyperparam_sensitivity}
\end{figure}

\begin{figure}[t]
\vspace{-1em}
\centering
\begin{subfigure}[t]{0.64\textwidth}
\centering
\includegraphics[width=\textwidth]{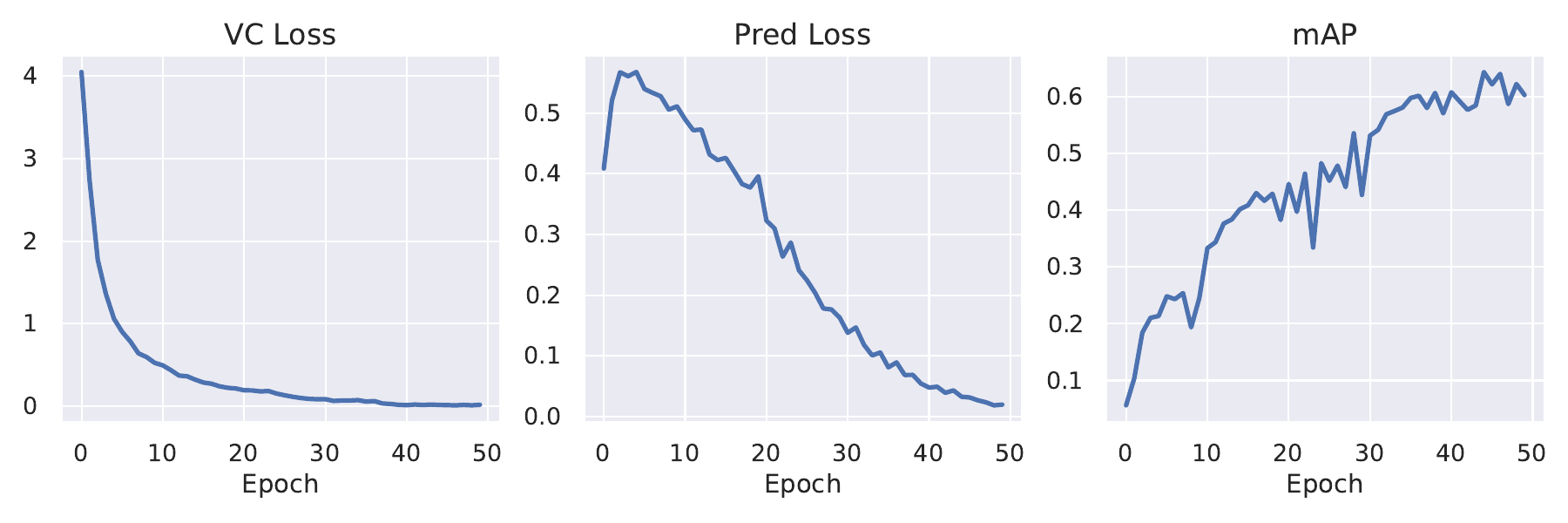}
\subcaption{}
\label{fig:multistep_pred_losses}
\end{subfigure}
\hfill
\begin{subfigure}[t]{0.35\textwidth}
\centering
\includegraphics[width=\textwidth]{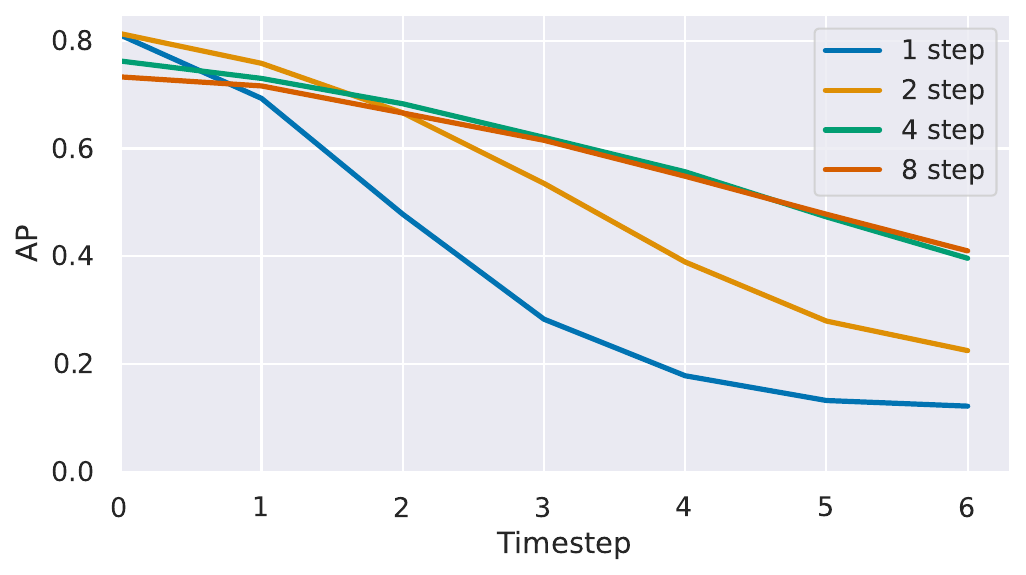}
\subcaption{}
\label{fig:multistep_pred_ap}
\end{subfigure}
\caption{Video-JEPA training dynamics and multistep rollout ablation. (a)~Training dynamics over 50 epochs: variance-covariance regularization loss $\mathcal{R}$ (left), prediction loss $\mathcal{L}_{\text{pred}}$ (center), and mean Average Precision (right). (b)~Training with $k$-step recursive predictions achieves significantly better Average Precision compared to single-step predictions, demonstrating improved temporal understanding, with a Pareto optimum around $k=4$ rollout steps.}
\label{fig:multistep_pred}
\end{figure}

\begin{figure}[t]
\centering
\includegraphics[width=\textwidth]{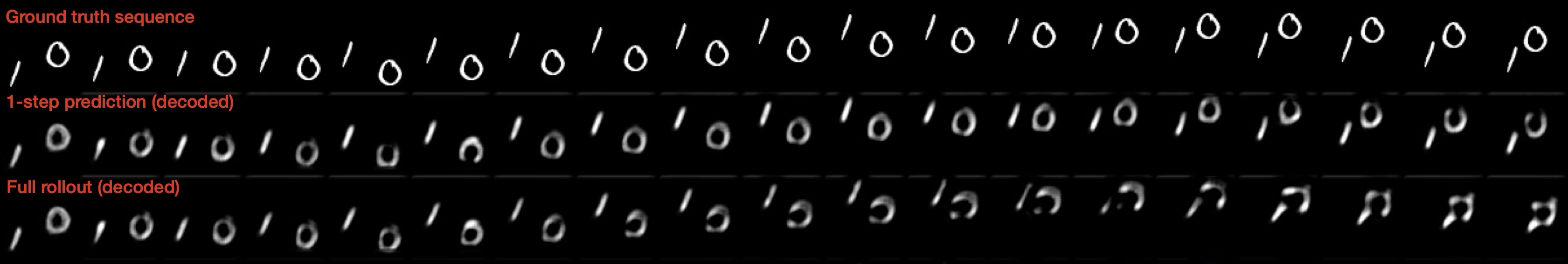}
\caption{Video JEPA visualization on Moving MNIST. From left to right: input frames, 1-step prediction visualization, and full autoregressive rollout. The model maintains coherent predictions of digit motion over extended horizons, correctly capturing trajectory and dynamics.}
\label{fig:video_viz}
\end{figure}

\section{Training and Planning with World Models}\label{sec:planning}

\paragraph{Multistep Rollout Training.}
In practice, for both video JEPA and Action-Conditioned JEPA, we augment single-step prediction with multistep rollout losses, following~\citet{terver2026JEPAWMs,VJEPA2}.
At each training iteration, we compute $k$-step rollout losses $\mathcal{L}_k$ for $k = 1, \dots, K$, where $\mathcal{L}_1$ recovers the single-step loss of Eqs.~\eqref{eq:video_jepa_energy}--\eqref{eq:ac_video_jepa_energy}.
Let us define the order of a prediction as the number of calls to the predictor function required to obtain it from a groundtruth representation.
For a predicted representation $z^{(k)}_t$, we denote the timestep it corresponds to as $t$ and its prediction order as $k$, with $z^{(0)}=z=f_{\theta}(x)$.
For $k \geq 1$, $\mathcal{L}_k$ is defined as
\begin{equation}\label{eq:multistep_rollout_loss}
    \mathcal{L}_k = \sum_{t=1}^{T-k} \| g_\phi(z^{(k-1)}_{t-v:t}, u_{t-v:t}) - z_{t+1} \|_2^2,
\end{equation}
where $z^{(k)}_{t}$ is obtained by recursively unrolling the predictor for all $t\leq T$, as
\begin{equation}\label{eq:unroll}
    z^{(k)}_{t+1} = g_\phi(z^{(k-1)}_{t-v:t}, u_{t-v:t}), \quad z^{(0)}_{t}=f_{\theta}(x_{t-w:t}).
\end{equation}
The total energy function losses of Eqs.~\eqref{eq:video_jepa_energy}--\eqref{eq:ac_video_jepa_energy} then read
\begin{equation}
\mathcal{L}_{\text{video}} = \mathcal{L}_{{\text{pred}}} + \lambda \mathcal{R}(z_{1:T}), \quad \mathcal{L}_{\text{world}} = \mathcal{L}_{{\text{pred}}}+ \lambda \mathcal{R}(z_{1:T}, u_{1:T}), \quad \mathcal{L}_{{\text{pred}}}=\sum_{k=1}^{K} \mathcal{L}_k.
\label{eq:full_multistep_losses_video}
\end{equation}
Note that we could perform truncated backpropagation through time (TBPTT)~\citep{TBPTT2002tuto}, detaching the gradient after each call to the predictor.
Training with $k$-step rollouts aligns the training procedure with autoregressive inference, reducing exposure bias and improving long-horizon prediction quality (see Figure~\ref{fig:multistep_pred}).

\paragraph{Additional Regularizers for World Models.}
Training action-conditioned JEPAs in randomized environments requires additional regularization beyond VICReg or SIGReg terms.
The temporal similarity loss $\mathcal{L}_{\text{sim}}$ encourages smooth representation trajectories along action sequences, and the inverse dynamics model (IDM) loss~\citep{curiositydrivenexplorationselfsupervisedprediction} $\mathcal{L}_{\text{IDM}}$ predicts actions from consecutive representations. These losses read
\begin{equation}
\mathcal{L}_{\text{sim}} = \sum_t \|z_t - z_{t+1}\|_2^2, \quad \mathcal{L}_{\text{IDM}} = \sum_t \|a_t - \text{MLP}(z_t, z_{t+1})\|_2^2.
\end{equation}
This term is critical for preventing collapse from spurious correlations in randomized environments~\citep{sobal2022jepaslowfeatures}.
The full training objective for action-conditioned video-JEPA combines prediction with all regularization terms and reads
\begin{equation}
\mathcal{L} = \mathcal{L}_{\text{pred}} + \alpha \mathcal{L}_{\text{var}} + \beta \mathcal{L}_{\text{cov}} + \delta \mathcal{L}_{\text{sim}} + \omega \mathcal{L}_{\text{IDM}}.
\label{eq:ac_video_full_loss}
\end{equation}

\paragraph{Goal-Conditioned Planning.}
We perform goal-conditioned planning by optimizing action sequences to reach a goal observation $x_g$.
This extends the energy function from Eq.~\eqref{eq:energy} to trajectories: rather than measuring prediction error for a single step, we accumulate the energy over an imagined rollout towards the goal as
\begin{equation}\label{eq:planning_energy}
E_{\text{plan}}(a_{0:H}; x_0, x_g) = \sum_{t=1}^{H} \| f_\theta(x_g) - \hat{z}_t \|_2, \quad \text{where } \hat{z}_{t+1} = g_\phi(\hat{z}_{t-v:t}, u_{t-v:t}), \quad \hat{z}_0 = f_\theta(x_0).
\end{equation}
Low energy corresponds to action sequences that successfully reach the goal; planning thus reduces to finding the minimum-energy trajectory.
We use Model Predictive Path Integral (MPPI)~\citep{MPPI}, a population-based optimizer that samples action trajectories, weights them by exponentiated negative energy (i.e., a Boltzmann distribution over trajectories), and iteratively refines the proposal distribution toward lower-energy solutions.
Summing over intermediate states (rather than only the final state) encourages efficient paths and provides robustness to prediction compounding errors.

\section{Experiments}

\paragraph{Experimental Setup.}
We evaluate the JEPA framework on three tasks of increasing complexity: image representation learning on CIFAR-10, video prediction on Moving MNIST~\citep{moving_mnist}, and goal-conditioned planning on the Two Rooms environment~\citep{PLDM}.
Our implementation uses modular building blocks:
\textbf{Encoders} (ResNet-18~\citep{ResNet}, Vision Transformers~\citep{VITs}, IMPALA~\citep{Impala}),
\textbf{Predictors} (UNet-based spatial predictors, GRU-based temporal predictors),
\textbf{Regularizers} (VICReg, SIGReg, temporal similarity, inverse dynamics losses), and
\textbf{Planners} (MPPI~\citep{MPPI} and Cross-Entropy Method (CEM) optimizers).
We provide comprehensive hyperparameter tables in Appendix~\ref{app:hyperparams}: Tables~\ref{tab:image_hyperparams} and~\ref{tab:video_hyperparams} summarize the best training hyperparameters for each example, and Table~\ref{tab:planning_hyperparams} details the planning configuration.

\paragraph{Image Representation Learning.} Tables~\ref{tab:image_results},~\ref{tab:image_results_params}, and~\ref{tab:image_results_proj} compare VICReg and SIGReg on CIFAR-10, using a naive hyperparameter search.
Both methods achieve approximately 90-91\% linear probing accuracy, competitive with prior self-supervised methods on this benchmark.
We find that using a learned projector provides around a 3 point improvement over directly regularizing encoder outputs. Projector architecture matters: a bottleneck design (large hidden $\rightarrow$ small output) works best for SIGReg, while VICReg prefers larger output dimensions. Having only one hyperparameter, SIGReg can be easier to tune in this naive setting (Figure~\ref{fig:hyperparam_sensitivity}).

\begin{table}[t]
\caption{Image-JEPA Linear probing accuracy on CIFAR-10 with ResNet-18 backbone trained for 300 epochs, comparing regularizers (SIGReg and VICReg) and the impact of using a projector.}
\label{tab:image_results}
\centering
\begin{tabular}{lccccc}
\toprule
          & Best acc. $\uparrow$ & Average acc. $\uparrow$ & w/o Projector & Hyperparams & Best projector \\
\midrule
SIGReg    & 91.02\%       & 89.22\%                           & -3.3 points       & 1               & 2048$\times$128     \\
VICReg    & 90.12\%       & 84.90\%                           & -2.9 points       & 2               & 2048$\times$1024    \\
\bottomrule
\end{tabular}
\end{table}

\begin{table}[t]
\caption{Ablation of Image-JEPA on loss hyperparameters when training on CIFAR-10 with ResNet-18 backbone trained for 300 epochs.}
\label{tab:image_results_params}
\centering
\begin{tabular}{lccccc}
\toprule
     & \multicolumn{2}{c}{SIGReg} & \multicolumn{2}{c}{VICReg} \\
\cmidrule(lr){2-3}\cmidrule(lr){4-5}
Rank & Hyperparameters & Accuracy $\uparrow$ & Hyperparameters & Accuracy $\uparrow$ \\
\midrule
1  & $\lambda = 10$   & 90.88\% & std = 1, cov = 100   & 90.12\% \\
2  & $\lambda = 1$    & 86.94\% & std = 1, cov = 10    & 89.93\% \\
3  & $\lambda = 100$  & 80.86\% & std = 10, cov = 10   & 89.20\% \\
-1 & $\lambda = 0.1$  & 27.20\% & std = 100, cov = 100 & 10.00\% \\
\bottomrule
\end{tabular}
\end{table}

\begin{table}[t]
\caption{Image-JEPA ablation of projector design when training on CIFAR-10 with ResNet-18 backbone trained for 300 epochs.}
\label{tab:image_results_proj}
\centering
\begin{tabular}{lcccc}
\toprule
     & \multicolumn{2}{c}{SIGReg} & \multicolumn{2}{c}{VICReg} \\
\cmidrule(lr){2-3}\cmidrule(lr){4-5}
Rank & Dimensions & Accuracy $\uparrow$ & Dimensions & Accuracy $\uparrow$ \\
\midrule
1   & 2048$\times$128  & 91.02\% & 2048$\times$1024 & 90.12\% \\
2   & 4096$\times$1024 & 91.00\% & 4096$\times$512  & 90.10\% \\
3   & 2048$\times$64   & 90.99\% & 1024$\times$1024 & 90.05\% \\
4   & 512$\times$256   & 90.99\% & 2048$\times$512  & 90.03\% \\
5   & 4096$\times$64   & 90.96\% & 4096$\times$1024 & 90.02\% \\
N/A & None             & 87.75\% & None             & 87.27\% \\
\bottomrule
\end{tabular}
\end{table}

\paragraph{Video Prediction.}
Multi-step autoregressive rollouts on Moving MNIST maintain prediction quality over extended horizons (Figure~\ref{fig:video_viz}).
Training with $k$-step prediction (rather than single-step) significantly improves Average Precision on downstream detection tasks by reducing exposure bias, i.e., the discrepancy between teacher-forced training and autoregressive inference.
Figure~\ref{fig:multistep_pred} shows that models trained with longer prediction horizons achieve better downstream performance, as recursive prediction during training aligns with the autoregressive inference procedure.

\begin{table}[t]
\caption{AC-video-JEPA planning ablations on Two Rooms with randomized wall positions. Each result averages over 3 seeds $\times$ 3 checkpoints $\times$ 20 episodes. \textbf{Left:} Planner configuration comparison. \textbf{Right:} Regularization term ablation; removing IDM causes collapse.}
\label{tab:planning_ablations}
\centering
\begin{minipage}[t]{0.48\textwidth}
\centering
\begin{tabular}{lcc}
\toprule
\textbf{Configuration} & \textbf{Success} $\uparrow$ & \textbf{Time} \\
\midrule
MPPI (full cost) & $97 \pm 2$\% & 37s \\
CEM (full cost) & $96 \pm 2$\% & 37s \\
MPPI (last state) & $89 \pm 2$\% & 37s \\
\bottomrule
\end{tabular}
\label{tab:planning_results}
\end{minipage}
\hfill
\begin{minipage}[t]{0.48\textwidth}
\centering
\begin{tabular}{lc}
\toprule
\textbf{Ablated Term} & \textbf{Success} $\uparrow$ \\
\midrule
None (full model) & $97 \pm 2$\% \\
Variance ($\alpha = 0$) & $47 \pm 3$\% \\
Covariance ($\beta = 0$) & $46 \pm 3$\% \\
Temporal Sim. ($\delta = 0$) & $61 \pm 2$\% \\
IDM ($\omega = 0$) & $1 \pm 1$\% \\
\bottomrule
\end{tabular}
\label{tab:ablation}
\end{minipage}
\end{table}

\begin{figure}[t]
\centering
\includegraphics[width=\textwidth, trim= 0pt 0pt 0pt 0pt]{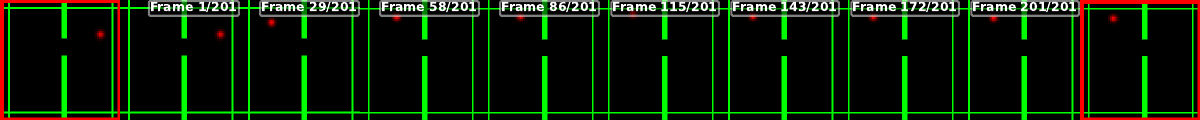}
\includegraphics[width=\textwidth, trim= 0pt 0pt 0pt 0pt]{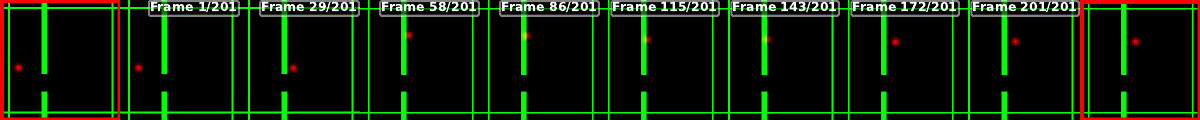}
\includegraphics[width=\textwidth, trim= 0pt 0pt 0pt 0pt]{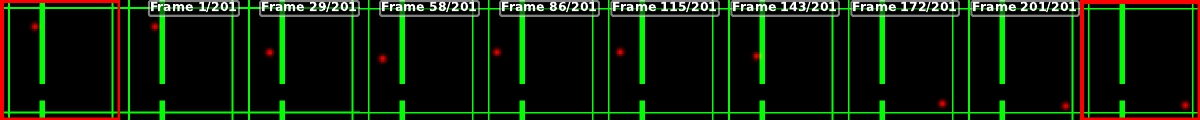}
\caption{Visualization of three successful planning evaluation episodes of our AC-video-JEPA on the Two Rooms environment with random wall. From left to right: initial frame (red), full episode outputted by the planning optimization procedure, goal frame used to define planning cost (red). Each episode allows a maximum of 200 steps in the environment.}
\label{fig:two_rooms_plan_episodes}
\vspace{-1em}
\end{figure}

\paragraph{Action-Conditioned Video-JEPA.}
We display three successful planning evaluation episodes in Figure~\ref{fig:two_rooms_plan_episodes}, showing the ability of the model to plan given randomized initial and goal states. This navigation task is non-monotonic, meaning that the optimal trajectory requires first getting farther from the goal, in order to reach it ultimately.
Table~\ref{tab:planning_results} shows planning results on the challenging random-wall setup.
Our best model achieves 97\% success rate using MPPI with cumulative cost over the planning horizon.

We perform an \textbf{ablation of the regularization components} of the action-conditioned video-JEPA models. Table~\ref{tab:ablation} reveals the importance of each regularization component: IDM is critical (without it, the model collapses to 1\% success due to spurious correlations~\citep{sobal2022jepaslowfeatures}); variance and covariance terms each contribute $\sim$50\% absolute improvement; temporal similarity adds $\sim$35\%.

We ablate the importance of \textbf{planning cost design}. Using cumulative cost over all timesteps ($\sum_t \|z_g - \hat{z}_t\|$) outperforms final-state-only cost by 8\% (Table~\ref{tab:planning_results}).
This formulation encourages efficient paths and provides gradient signal throughout the trajectory.

\section{Future Directions}

EB-JEPA is designed for fast iteration on algorithmic innovations at small scale: single-GPU training, simple datasets, and controlled simulated environments.
This enables rapid prototyping and fundamental research on JEPA architectures before scaling to more complex settings.
We identify three promising algorithmic directions that EB-JEPA's modular design enables researchers to explore.

\paragraph{Advancing Regularization Theory.}
Our experiments highlight the critical role of regularization in preventing representation collapse, yet the theoretical understanding of why certain regularization combinations work remains incomplete.
EB-JEPA provides a testbed for systematically studying regularization dynamics: investigating the interplay between variance, covariance, temporal similarity, and inverse dynamics terms~\citep{VICReg,balestriero2025lejepaprovablescalableselfsupervised,sobal2022jepaslowfeatures}; understanding when each becomes necessary; and developing principled methods for automatic hyperparameter selection.
The controlled, single-GPU setting enables rapid iteration on these fundamental questions without the confounding factors introduced by large-scale distributed training.

\paragraph{Hierarchical World Models.}
Current JEPA models predict at a single temporal resolution, but intelligent planning often requires reasoning at multiple timescales~\citep{Schmidhuber2015,Director}.
Hierarchical world models could learn to predict both fine-grained dynamics for local control and coarse-grained abstractions for long-horizon planning.
Prior work in hierarchical reinforcement learning~\citep{HIRO,HAC} has demonstrated the benefits of learning at multiple levels of abstraction.
EB-JEPA's separation of encoder, predictor, and regularizer components provides a natural starting point for implementing such multi-scale architectures, and future releases may include basic hierarchical prediction examples.

\paragraph{Learned Cost and Value Functions.}
Our current planning approach uses simple distance-based costs in representation space, but this may be suboptimal for complex tasks.
Learning task-specific cost functions or value functions from demonstrations or sparse rewards could enable more sophisticated goal-directed behavior.
Combining JEPA world models with learned value functions~\citep{TD-MPC,TDMPC2} offers a promising avenue for making better use of the predictive models trained with this codebase, potentially bridging the gap between pure world modeling and reward-driven reinforcement learning.
EB-JEPA's simple planning interface makes it straightforward to experiment with alternative cost formulations.

\paragraph{Complementary to Large-Scale Codebases.}
EB-JEPA is intended for algorithmic exploration and fundamental research.
Once promising approaches are validated at small scale, researchers can transition to codebases supporting distributed training, pre-trained visual backbones, and more complex environments, such as JEPA-WMs~\citep{terver2026JEPAWMs} for planning with frozen encoders on diverse benchmarks, and \texttt{stable-pretraining}~\citep{balestriero2025stablepretrainingv1} for general self-supervised learning (SSL) pretraining.
EB-JEPA complements these by prioritizing educational simplicity and self-contained, single-GPU examples.
This two-stage workflow enables efficient research: rapid prototyping with EB-JEPA followed by rigorous evaluation at scale.

\section{Conclusion}

We have presented EB-JEPA, an open-source library for learning representations and world models using Joint-Embedding Predictive Architectures.
Our implementations span image representation learning, video prediction, and action-conditioned planning, each designed to be trainable on a single GPU within a few hours.
Comprehensive experiments demonstrate that our implementations achieve strong results on established benchmarks while providing insights into the importance of each component.
The ablation studies reveal that all regularization terms (variance, covariance, temporal similarity, and inverse dynamics) play important roles in preventing collapse and enabling effective planning.
We hope EB-JEPA serves as both a practical toolkit for researchers exploring JEPA-based methods and an educational resource for understanding energy-based self-supervised learning.

\newpage

\subsection*{Ethics statement}

EB-JEPA is an educational library for self-supervised learning research. All experiments use standard public benchmarks (CIFAR-10, Moving MNIST) or procedurally generated environments (Two Rooms). None of these datasets contain personally identifiable information. We see no direct ethical concerns with this work.

\subsection*{Reproducibility statement}

Reproducibility is the central goal of this work. Our full codebase is included in the supplementary material, with all training scripts, model implementations, and evaluation code. Each example is self-contained and trains on a single GPU in a few hours, removing the need for large compute clusters. Hyperparameters for all experiments are listed in Appendix~\ref{app:hyperparams}. The Two Rooms environment is procedurally generated with documented seeds.

\ificlrfinal
\subsubsection*{Acknowledgments}
We thank Adrien Bardes and Gaoyue Zhou for participating in the discussions and conceptualization of the project.
\fi

\newpage
\bibliography{main}
\bibliographystyle{iclr2026_conference}

\appendix
\newpage

\section{Hyperparameters}\label{app:hyperparams}

This section provides the hyperparameters used for training and evaluation across our examples.
Tables~\ref{tab:image_hyperparams} and~\ref{tab:video_hyperparams} summarize the key training hyperparameters, including the number of rollout steps $K$ used for multistep prediction training (Eq.~\eqref{eq:multistep_rollout_loss}) and the trajectory slice length $T$ for temporal examples.
Table~\ref{tab:planning_hyperparams} details the MPPI planning configuration used for goal-conditioned navigation in the action-conditioned video-JEPA example.

\begin{table}[h]
\centering
\caption{Training hyperparameters for image-JEPA examples on CIFAR-10. The ``ViT Image-JEPA'' column documents hyperparameters for a ViT-based example provided in the codebase, which achieves 87\% linear probing accuracy; all image results reported in the main paper (Tables~\ref{tab:image_results},~\ref{tab:image_results_params}, and~\ref{tab:image_results_proj}) use the ResNet-18 backbone.}
\label{tab:image_hyperparams}
\begin{tabular}{llccc}
\toprule
\textbf{Group} & \textbf{Hyperparameter} & \textbf{VICReg} & \textbf{ViT Image-JEPA} & \textbf{SIGReg} \\
\midrule
\multirow{6}{*}{Optimization} & Optimizer & LARS & AdamW & LARS \\
 & LR schedule & \multicolumn{3}{c}{10-epoch warmup (from $3 \times 10^{-5}$) + cosine to 0} \\
 & Learning rate & 0.3 & 0.001 & 0.3 \\
 & Epochs & 300 & 300 & 300 \\
 & Batch size & 256 & 512 & 256 \\
 & Weight decay & $10^{-4}$ & $10^{-4}$ & $10^{-4}$ \\
\midrule
\multirow{2}{*}{Data} & Dataset & CIFAR-10 & CIFAR-10 & CIFAR-10 \\
 & Image size & $32 \times 32$ & $32 \times 32$ & $32 \times 32$ \\
\midrule
\multirow{6}{*}{Architecture} & Encoder & ResNet-18 & ViT-S & ResNet-18 \\
 & Predictor & Identity & Identity & Identity \\
 & Encoder output dim & 512 & 384 & 512 \\
 & Projector hidden dim & 2048 & 2048 & 2048 \\
 & Projector output dim & 2048 & 2048 & 128 \\
 & Projector layers & 3 & 3 & 3 \\
\midrule
\multirow{4}{*}{Loss} & Loss type & VICReg & VICReg & SIGReg \\
 & Variance coeff. $\alpha$ & 1 & 1 & -- \\
 & Covariance coeff. $\beta$ & 80 & 80 & -- \\
 & SIGReg coeff. $\lambda$ & -- & -- & 10 \\
\bottomrule
\end{tabular}
\end{table}

\begin{table}[h]
\centering
\caption{Training hyperparameters for video-JEPA examples. $K$ denotes the number of training rollout steps (multistep prediction), and $T$ denotes the training trajectory slice length.}
\label{tab:video_hyperparams}
\begin{tabular}{llcc}
\toprule
\textbf{Group} & \textbf{Hyperparameter} & \textbf{Video-JEPA} & \textbf{AC-Video-JEPA} \\
\midrule
\multirow{4}{*}{Optimization} & Learning rate & 0.001 & 0.001 \\
 & Epochs & 50 & 12 \\
 & Batch size & 64 & 384 \\
 & Weight decay & -- & $10^{-5}$ \\
\midrule
\multirow{3}{*}{Data} & Dataset & Moving MNIST & Two Rooms \\
 & Trajectory length $T$ & 10 & 17 \\
 & Image size & $64 \times 64$ & $65 \times 65$ \\
\midrule
\multirow{6}{*}{Architecture} & Encoder & ResNet-5 & IMPALA \\
 & Predictor & ResUNet & GRU \\
 & Latent dimension $d$ & 16 & 32 \\
 & Hidden dimension & 32 & 32 \\
 & Encoder receptive field $w$ & 1 & 1 \\
 & Predictor receptive field $v$ & 2 & 1 \\
\midrule
\multirow{5}{*}{Loss} & Rollout steps $K$ & 4 & 8 \\
 & Variance coeff. $\alpha$ & 10 & 16 \\
 & Covariance coeff. $\beta$ & 100 & 8 \\
 & Time similarity coeff. $\delta$ & -- & 12 \\
 & IDM coeff. $\omega$ & -- & 1 \\
\bottomrule
\end{tabular}
\end{table}

\begin{table}[h]
\centering
\caption{Planning hyperparameters for the action-conditioned video-JEPA example using MPPI, corresponding to the notations of Algorithm~\ref{algo:MPPI}. The total number of replanning steps for an evaluation episode is $\frac{M}{m}$.
}
\label{tab:planning_hyperparams}
\begin{tabular}{llc}
\toprule
\textbf{Hyperparameter} & \textbf{Symbol} & \textbf{Value} \\
\midrule
Planning horizon & $H$ & 90 \\
Number of parallel samples & $N$ & 200 \\
Number of iterations & $J$ & 20 \\
Number of elites & $Q$ & 20 \\
Noise scale & $\sigma$ & 2 \\
Temperature & $\tau$ & 0.005 \\
Actions stepped per plan & $m$ & 1 \\
Max steps per episode & $M$ & 200 \\
\bottomrule
\end{tabular}
\end{table}

\section{Planning Algorithm}\label{app:planning_algo}

We use MPPI control~\citep{MPPI} for planning, a sampling-based optimization algorithm that uses importance sampling to iteratively refine action sequences.
Unlike the CEM which fits a Gaussian to elite samples, MPPI weights all samples by their exponentiated costs, providing smoother gradient information and better exploration.

Given a trained encoder $f_\theta$, predictor $g_\phi$, and action encoder $q_\omega$, we minimize the planning energy $E_{\text{plan}}$ from Eq.~\eqref{eq:planning_energy} over action sequences as described in Algorithm~\ref{algo:MPPI}.

\begin{algorithm}
\caption{Model Predictive Path Integral (MPPI)}
\label{algo:MPPI}
\begin{algorithmic}[1]
    \STATE \textbf{Input:} Initial observation $x_0$, goal observation $x_g$, initial mean $\mu \in \mathbb{R}^{H \times A}$, noise scale $\sigma$, temperature $\tau$, number of samples $N$, number of iterations $J$, number of elites $Q$, max steps per episode $M$
    \STATE Encode initial and goal: $\hat{z}_0 = f_\theta(x_0)$, $z_g = f_\theta(x_g)$
    \FOR{$j = 1$ to $J$}
        \STATE Sample $N$ noise perturbations: $\epsilon_i \sim \mathcal{N}(0, \sigma^2 \textrm{I})$ for $i = 1, \ldots, N$
        \STATE Compute candidate action sequences: $a^{(i)}_{0:H-1} = \mu + \epsilon_i$
        \STATE Unroll predictor for each trajectory: $\hat{z}^{(i)}_{t+1} = g_\phi(\hat{z}^{(i)}_{t-v:t}, u^{(i)}_{t-v:t})$ for $t = 0, \ldots, H-1$
        \STATE Compute trajectory costs: $S_i = \sum_{t=1}^{H} \| z_g - \hat{z}^{(i)}_t \|_2$
        \STATE Select top $Q$ elite samples with lowest costs
        \STATE Compute weights over elites: $w_i = \frac{\exp(-S_i / \tau)}{\sum_{q=1}^{Q} \exp(-S_q / \tau)}$
        \STATE Update mean: $\mu \leftarrow \sum_{i=1}^{Q} w_i \cdot a^{(i)}_{0:H-1}$
    \ENDFOR
    \STATE \textbf{Return:} Execute first $m$ actions of $\mu$, then replan from new observation until $M$ steps reached
\end{algorithmic}
\end{algorithm}

The key differences from CEM are: (1) MPPI uses soft weighting via the exponential transform rather than hard elite selection for the update, (2) the temperature parameter $\tau$ controls the sharpness of the weight distribution, and (3) MPPI naturally handles multi-modal cost landscapes through its importance sampling formulation.
In our implementation, we combine MPPI with elite selection: we first select the top $Q$ trajectories, then apply exponential weighting only among these elites, which provides both the robustness of elite selection and the smooth gradients of importance weighting.

\section{Extended Related Work}

\paragraph{Diffusion-Based Planning.}
An alternative paradigm for planning uses diffusion models to generate trajectories.
Diffuser~\citep{Diffuser} pioneered planning with diffusion by treating trajectory optimization as iterative denoising.
Diffusion MPC~\citep{DMPC} extends this to model predictive control settings, while Diffusion Policy~\citep{chi2023diffusion} applies diffusion to visuomotor policy learning.
These approaches complement JEPA-based methods: while diffusion models excel at generating diverse, multimodal trajectories, JEPAs provide efficient latent dynamics suitable for fast online planning.

\end{document}